# FAULT CLASSIFICATION IN CYLINDERS USING MULTI-LAYER PERCEPTRONS, SUPPORT VECTOR MACHINES AND GUASSIAN MIXTURE MODELS

**Tshilidzi Marwala[a], Unathi Mahola[a] and Snehashish Chakraverty[b]**

*[a]School of Electrical and Information Engineering*
*University of the Witwatersrand Private Bag x 3 Wits 2050 South Africa*
*e-mail: t.marwala@ee.wits.ac.za*
*[b]Central Building Research Institute*
*Roorkee-247 667, U.A. India*
*e-mail: sne_chak@yahoo.com*

**Abstract**

Gaussian mixture models (GMM) and support vector machines (SVM) are introduced to classify faults in a population of cylindrical shells. The proposed procedures are tested on a population of 20 cylindrical shells and their performance is compared to the procedure, which uses multi-layer perceptrons (MLP). The modal properties extracted from vibration data are used to train the GMM, SVM and MLP. It is observed that the GMM produces 98%, SVM produces 94% classification accuracy while the MLP produces 88% classification rates.

## 1. INTRODUCTION

Vibration data have been used with varying degrees of success to classify damage in structures [1]. In the fault classification process there are various stages involved and these are: data extraction, data processing, data analysis and fault classification. Data extraction process involves the choice of data to be extracted and the method of extraction. Data that have been used for fault classification include strains concentration in structures and vibration data where strain gauges and accelerometers are used respectively [1]. In this paper vibration data processed using modal analysis are used for fault classification.

In the data processing stage the measured vibration data need to be processed. This is mainly due to the fact that the measured vibration data, which are in the time domain, are difficult to use in raw form. Thus far the time-domain vibration data may be transformed to the modal analysis, frequency domain analysis and time-frequency domain [2,3]. In this paper the time-domain vibration data set is transformed into the modal domain where it is represented as natural frequencies and mode shapes.

The data processed need to be analysed and the general trend has been to automate the analysis process and thus automate the fault classification process. To achieve this goal intelligent pattern recognition process needs to be employed and methods such as neural networks have been widely applied [1]. There are many types of neural networks that have been employed and these include multi-layer perceptron (MLP), radial basis function (RBF) and Bayesian neural networks [4,5]. Recently, new pattern recognition methods called support vector machines (SVMs) and Gaussian mixture models (GMMs) have been proposed and found to be particularly suited to classification problems [6]. SVMs have been found to outperform neural networks [7]. One of the examples where the fault classification process summarized at the beginning of this paper has been implemented is fault classification in a population of nominally identical cylindrical shells [2,3,4]. The importance of fault identification process in a population of nominally identical structures is particularly important in areas such as the automated manufacturing process in the assembly line. Thus far various forms of neural networks such as MLP and Bayesian neural networks have been successfully used to classify faults in structures [8]. Worden and Lane [9] used SVMs to identify damage in structures. However, SVMs have not been used for fault classification in a population of cylinders. Based on the successes of SVMs observed in other areas, we therefore propose in this paper SVMs and GMMs for classifying faults in a population of nominally identical cylindrical shells.

This paper is organized as follows: neural networks, SVMs and GMMs are summarized, experiment performed is discussed and results as well as conclusions are discussed.

## 2. NEURAL NETWORKS

Neural networks are parameterised graphs that make probabilistic assumptions about data and in this paper these data are modal domain data and their respective classes of faults. In this paper multi-layer perceptron neural networks are trained to give a relationship between modal domain data and the fault classes.

As mentioned earlier, there are several types of neural network procedures such as multi-layer perceptron, radial basis function, Bayesian networks and recurrent networks [5] and in this paper the MLP is used.

This network architecture contains hidden units and output units and has one hidden layer. For the MLP the relationship between the output y, representing fault class, and x, representing modal data, may be may be written as follows [5,10,11]:

$$y_k = f_{outer}\left(\sum_{j=1}^{M} w_{kj}^{(2)} f_{inner}\left(\sum_{i=1}^{d} w_{ji}^{(1)} x_i + w_{j0}^{(1)}\right) + w_{k0}^{(2)}\right) \quad (1)$$

Here, $w_{ji}^{(1)}$ and $w_{ji}^{(2)}$ indicate weights in the first and second layer, respectively, going from input i to hidden unit j, M is the number of hidden units, d is the number of output units while $w_{j0}^{(1)}$ and $w_{k0}^{(2)}$ indicate the biases for the hidden unit j and the output unit k. In this paper, the function $f_{outer}(\bullet)$ is logistic while $f_{inner}$ is a hyperbolic tangent function. Training the neural network identifies the weights in equations 1 and in this paper the scaled conjugate gradient method is used [12].

## 3. SUPPORT VECTOR MACHINES (SVMs)

According to [13], the classification problem can be formally stated as estimating a function f: RN → {−1, 1} based on an input-output training data generated from an independently, identically distributed unknown probability distribution P(x, y) such that f will be able to classify previously unseen (x, y) pairs. Here x is the modal data while y is the fault class. The best such function is the one that minimizes the expected error (risk) which is given by

$$R[f] = \int l(f(x), y)\,dP(x, y) \quad (2)$$

where l represents a loss function [13]. Since the underlying probability distribution P is unknown, equation 2 cannot be solved directly. The best we can do is finding an upper bound for the risk function [14] that is given by

$$R[f] = R[f]_{emp} + \sqrt{\frac{h(\ln\frac{2n}{h}+1) - \ln(\delta/4))}{n}} \quad (3)$$

where h∈ $N^+$ is the Vapnik-Chervonenkis (VC) dimension of f∈ F and δ > 0 holds true for all δ. The VC dimension of a function class F is defined as the largest number h of points that can be separated in all possible ways using functions of the class [14]. The empirical error $R[f]_{emp}$ is a training error given by

$$R[f]_{emp} = \frac{1}{n}\sum_{i+1}^{n} l(f(x_i), y_i) \quad (4)$$

Assuming that the training sample is linearly separable by a hyperplane of the form

$$f(x) = \langle w, x\rangle + b \quad \text{with} \quad w \in \chi \ , \ b \in \Re \quad (5)$$

where w is an adjustable weight vector and b is an offset, the classification problem looks like Figure 1 [14].

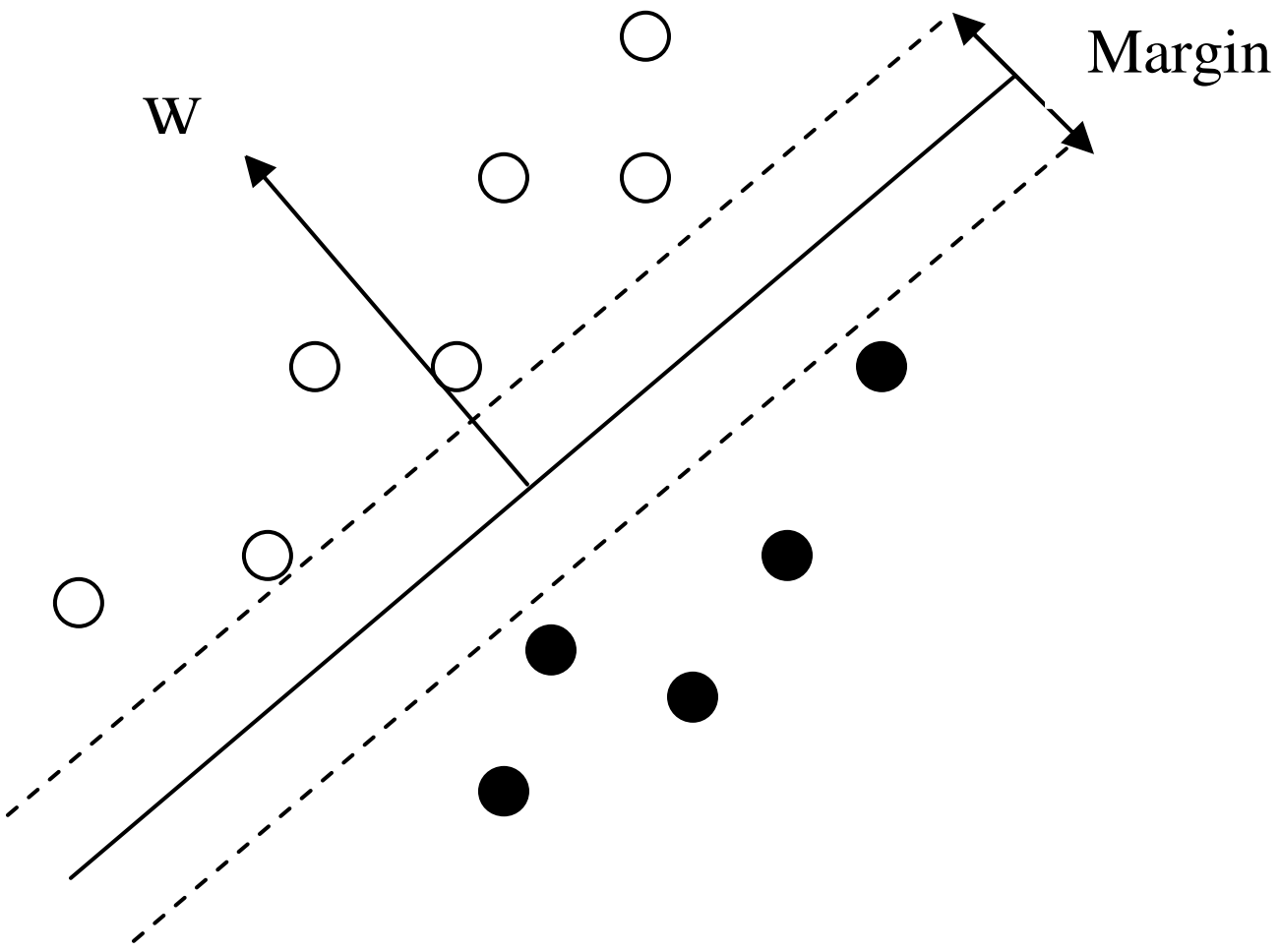

**Figure 1. A linear SVM classifier and margins: A linear classifier is defined by a hyperplane's normal vector w and an offset b, i.e. the decision boundary is {x|w.x + b =0} (thick line). Each of the two half spaces defined by this hyperplane corresponds to one class, i.e. f(x) = sign((w.x) + b).[13]**

The goal of the learning algorithm as proposed by [12], is to find the hyperplane with maximum margin of separation from the class of separating hyperplanes. But since real-world data often exhibit complex properties, which cannot be separated linearly, more complex classifiers are required. In order to avoid the complexity of the nonlinear classifiers, the idea of linear classifiers in a feature space comes into place. Support vector machines try to find a linear separating hyperplane by first mapping the input space into a higher dimensional feature space F. This implies each training example $x_i$ is substituted with $\Phi(x_i)$.

$$y_i((w.\phi(x_i) + b), i = 1,2,...,n \quad (6)$$

The VC dimension h in the feature space F is bounded according to $h \le \|W\|^2 R^2 + 1$ where R is the radius of the smallest sphere around the training data [12]. Hence minimizing the expected risk is stated as an optimisation problem

$$\min_{w,b} \ \frac{1}{2}\|w\|^2 \quad (7)$$

However, assuming that we can only access the feature space using only dot products, equation 7 is transformed into a dual optimization problem by introducing Lagrangian multipliers $\alpha_i, i = 1, 2, ..., n$ and doing some minimisation, maximisation and saddle point property of the optimal point [14,15,16] the problem becomes

$$\max_{w,b} \sum_{i=1}^{n} \alpha_i - \frac{1}{2}\sum_{i,j=1}^{n} \alpha_i \alpha_j y_i y_j k(x_i, x_j)$$

$$\text{subject to} \quad \alpha_i \geq 0, \ i = 1,...,n \quad \sum_{i=1}^{n} \alpha_i y_i = 0 \tag{8}$$

The Lagrangian coefficients, $\alpha_i$'s, are obtained by solving equation 8 which in turn is used to solve w to give the non-linear decision function [12]

$$f(x) = \operatorname{sgn}\left(\sum_{i=1}^{n} y_i \alpha_i \left(\Phi(x).\Phi(x_i)\right) + b\right) = \operatorname{sgn}(\sum_{i=1}^{n} y_i \alpha_i k(x, x_i) + b \tag{9}$$

In the case when the data is not linearly separable, a slack variable $\xi_i$, i = 1,..., n is introduced to relax the constraints of the margin as

$$y_i((w.\Phi(x_i)) + b) \geq 1 - \zeta_i, \ \zeta_i \geq 0, \ i = 1,...,n \tag{10}$$

A trade off is made between the VC dimension and the complexity term of equation 3, which gives the optimisation problem

$$\min_{w,b,\xi} \quad \frac{1}{2}\|w\|^2 + C\sum_{i=1}^{1} \xi_i \tag{11}$$

where C > 0 is a regularisation constant that determines the above-mentioned trade-off. The dual optimisation problem is then given by [12]

$$\max_{\alpha} \quad \sum_{i=1}^{n} \alpha_i - \frac{1}{2}\sum_{i,j=1}^{n} \alpha_i \alpha_j y_i y_j k(x_i, x_j) \tag{12}$$

subject to

$$0 \leq \alpha_i \leq C, i = 1,...,n.$$

$$\sum_{i=1}^{n} \alpha_i y_i = 0$$

A Karush-Kuhn-Tucker (KKT) condition which says that only the $\alpha_i$'s associated with the training values $x_i$'s on or inside the *margin* area have non-zero values, is applied to the above optimisation problem to find the $\alpha_i$'s and the threshold variable b reasonably and the decision function f [17].

## 4. GAUSSIAN MIXTURE MODELS (GMMs)

GMM non-linear pattern classifier also works by creating a maximum likelihood model for each fault case given by [18],

$$\lambda = \{ w, \mu, \Sigma \} \tag{13}$$

where, w, $\mu$, $\Sigma$ are the weights, means and diagonal covariance of the features. Given a collection of training vectors the parameters of this model are estimated by a number of algorithms such as the Expectation-Maximization (EM) algorithm [18]. In this paper, the EM algorithm is used since it has reasonable fast computational time when compared to other algorithms. The EM algorithm finds the optimum model parameters by iteratively refining GMM parameters to increase the likelihood of the estimated model for the given fault feature modal vector. For the EM equations for training a GMM, the reader is referred to [19]. Fault detection or diagnosis using this classifier is then achieved by computing the likelihood of the unknown modal data of the different fault models. This likelihood is given by [18]

$$\hat{s} = \arg\max_{1 \le f \le F} \sum_{k=1}^{K} \log p(x_k \mid \lambda_f) \tag{14}$$

where, F, represent the number of faults to be diagonalized, $X = \{x_1, x_2, ..., x_K\}$ is the unknown D-dimension fault modal data and $p(x_k|\lambda_f)$ is the mixture density function given by [18]

$$p(x \mid \lambda) = \sum_{i=1}^{M} w_i p(x) \tag{15}$$

with,

$$p_i(x_t) = \frac{1}{(2\pi)^{D/2}\sqrt{\sum_i}} \exp\{-\frac{1}{2}(x_k - \mu_i)^T (\sum_i)^{-1}(x_k - \mu_i)\} \tag{16}$$

It should be noted that the mixture weights, $w_i$, satisfy the constrains, $\sum_{i=1}^{M} w_i = 1$.

## 5. INPUT DATA

This section describes the inputs that are used to test the SVM, MLP and GMM. When modal analysis is used for fault classification it is often found that there are more parameters extracted from the vibration data than can be possibly used for MLP, SVM and GMM training. These data must therefore be reduced because of a phenomenon called the curse of dimensionality [18], which refers to the difficulties associated with the feasibility of density estimation in many dimensions. However, this reduction process must be conducted such that the loss of essential information is minimized. The techniques implemented in this paper to reduce the dimension of the input data remove parts of the data that do not contribute significantly to the dynamics of the system being analysed or those that are too sensitive to irrelevant parameters. To achieve this, we implement the principal component analysis, which is discussed in the next section.

### 5.1 Principal Component Analysis

In this paper we use the principal component analysis (PCA) [20;21] to reduce the input data into independent input data. The PCA orthogonalizes the components of the input vector so that they are uncorrelated with each other. In the PCA, correlations and interactions among variables in the data are summarised in terms of a small number of underlying factors.

## 6. FOUNDATIONS OF DYNAMICS

As indicated earlier, in this paper modal properties i.e. natural frequencies and mode shapes are extracted from the measured vibration data and used for fault classification. For this reason the foundation of these parameters are described in this section. All elastic structures may be described the time domain as [22]

$$[M]\{X''\}+[C]\{X'\}+[K]\}\{X\}=\{F\} \quad (17)$$

where [M], [C] and [K] are the mass, damping and stiffness matrices respectively, and {X}, {X′} and {X″} are the displacement, velocity and acceleration vectors, respectively, while {F} is the applied force vector. If equation 17 is transformed into the modal domain to form an eigenvalue equation for the $i^{th}$ mode, then [23]

$$(-\bar{\omega}_i^2[M]+j\bar{\omega}_i[C]+[K])\{\bar{\phi}\}_i=\{0\} \quad (18)$$

where $j=\sqrt{-1}$, $\bar{\omega}_i$ is the $i^{th}$ complex eigenvalue, with its imaginary part corresponding to the natural frequency $\omega_i$, {0} is the null vector, and $\{\bar{\phi}\}_i$ is the $i^{th}$ complex mode shape vector with the real part corresponding to the normalized mode shape $\{\phi\}_i$. From equation 18 it is evident that changes in the mass and stiffness matrices cause changes in the modal properties and thus modal properties are damage indicators.

## 7. EXAMPLE: CYLINDRICAL SHELLS

### 7.1 Experimental Procedure

In this section the procedures of using GMM and SVM are experimentally validated and compared to the procedure of using MLP. The experiment is performed on a population of cylinders, which are supported by inserting a sponge rested on a bubble-wrap, to simulate a ‘free-free’ environment [see Figure 2]. The impulse hammer test is performed on each of the 20 steel seam-welded cylindrical shells. The impulse is applied at 19 different locations as indicated in Figure 2. More details on this experiment may be found in [4].

Each cylinder is divided into three equal substructures and holes of 10-15 mm in diameter are introduced at the centers of the substructures to simulate faults.

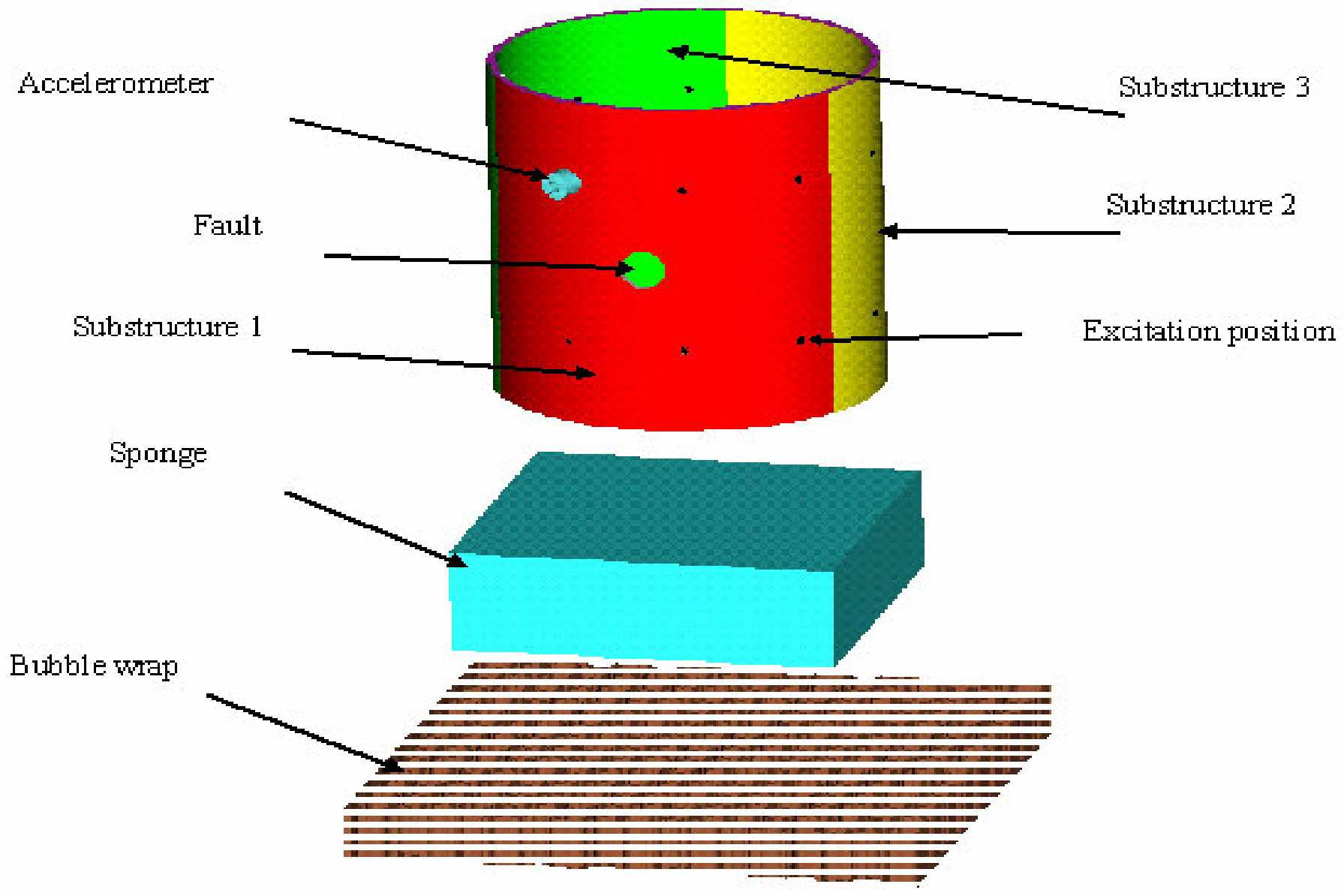

**Figure 2: Illustration of a cylindrical shell showing the positions of the impulse, accelerometer, substructures, fault position and supporting sponge.**

For one cylinder the first type of fault is a zero-fault scenario. This type of fault is given the identity [0 0 0], indicating that there are no faults in any of the three substructures. The second type of fault is a one-fault-scenario, where a hole may be located in any of the three substructures. Three possible one-fault-scenarios are [1 0 0], [0 1 0], and [0 0 1] indicating one hole in substructures 1, 2 or 3, respectively. The third type of fault is a two-fault scenario, where holes are located in two of the three substructures. Three possible two-fault-scenarios are [1 1 0], [1 0 1], and [0 1 1]. The final type of fault is a three-fault-scenario, where a hole is located in all three substructures, and the identity of this fault is [1 1 1]. There are 8 different types of fault-cases considered (including [0 0 0]).

Each cylinder is measured three times under different boundary conditions by changing the orientation of a rectangular sponge inserted inside the cylinder. The number of sets of measurements taken for undamaged population is 60 (20 cylinders × 3 for different boundary conditions). In the 8 possible fault types, two fault types [0 0 0] and [1 1 1] has 60 number of occurrences while the rest has 24. It should be noted that the numbers of one- and two-fault cases are each 72. This is because as mentioned above, increasing the sizes of holes in the substructures and taking vibration measurements generated additional one- and two-fault cases. Fault cases used to train and test the networks are shown in Table 1.

Table 1.Fault cases used to train, cross-validates and test the networks. The multifold cross-validation technique is used because of the lack of availability of the data

| Fault | [000] | [100] | [010] | [001] | [110] | [101] | [011] | [111] |
|---|---|---|---|---|---|---|---|---|
| Training set | 21 | 21 | 21 | 21 | 21 | 21 | 21 | 21 |
| Test set | 39 | 3 | 3 | 3 | 3 | 3 | 3 | 39 |

The impulse and response data are processed using the Fast Fourier Transform (FFT) to convert the time domain impulse history and response data into the frequency domain. The data in the frequency domain are used to calculate the frequency response functions (FRFs). From the FRFs, the modal properties are extracted using modal analysis [21]. The number of modal properties identified is 340 (17 modes×19 measured mode-shape-co-ordinates+17 natural frequencies). The PCA are used to reduce the dimension of the input data from 340×264 modal properties to 10×264. Here 264 correspond to the number of fault cases measured.

## 8. RESULTS AND DISCUSSION

The measured data was used for the MLP training and the MLP architecture contained 10 input units, 8 hidden units and 3 output units. The scaled conjugate gradient method was used for training the MLP network [12]. The average time it took to train the MLP networks was 12 CPU seconds on the Pentium II computer. The results obtained are shown in Table 2. In this table the actual fault cases are listed against the predicted fault cases. These results show that the MLP classify fault cases to the accuracy of 88%. In Table 1 it was shown that some fault cases are more numerous than others. In this case the measure of accuracy as a ratio of the sum of fault cases classified correctly divided by the total number of cases can be misleading. This is the case if the fault cases classified incorrectly are those from the less numerous cases. To remedy this situation a measure of accuracy called geometrical accuracy (GA) is used and defined as:

$$GA = \sqrt{\frac{\prod c_1 ... c_n}{\prod q_1 ... q_n}} \quad (19)$$

where $\prod$ is the product, $c_n$ is the number of $n^{th}$ fault cases classified correctly while $q_n$ is the $n^{th}$ fault class. Using this measure the MLP gives the accuracy of 0.7.On training the SVM, there are different parameters that can be changed namely the capacity and the e-insensitivity, the amount of training inputs and the function to be used for the kernel. Some of the functions that can be used are: linear, radial basis function, sigmoid, and spline. In this paper the exponential radial basis function is used as a kernel. The training process took 45 CPU seconds and the capacity was set to infinity. The results obtained are shown in Table 3 and these results show that the SVM gives accuracy of 94% while it gives the GA of 0.92. GMM architecture on the other hand used diagonal covariance matrix with 3 centers. The main advantage of using the diagonal covariance matrix is that this de-correlates the feature vectors. The training process took 45 CPU seconds. Table 4 shows that the GMM gives accuracy of 98% while it gives the GA of 0.95. As can be seen from Tables 2, 3 and 4, the GMM outperforms, the SVM network which out-performed the MLP. The MLP network.

Table 2. The confusion matrix obtained when the MLP network is used for fault classification

Predicted

| | | [000] | [100] | [010] | [001] | [110] | [101] | [011] | [111] |
|---|---|---|---|---|---|---|---|---|---|
| Actual | [000] | 39 | 0 | 0 | 0 | 0 | 0 | 0 | 0 |
| | [100] | 0 | 3 | 0 | 0 | 0 | 0 | 0 | 0 |
| | [010] | 0 | 0 | 3 | 0 | 0 | 0 | 0 | 0 |
| | [001] | 0 | 0 | 0 | 3 | 0 | 0 | 0 | 6 |
| | [110] | 0 | 0 | 0 | 0 | 3 | 1 | 0 | 0 |
| | [101] | 0 | 0 | 0 | 0 | 0 | 2 | 0 | 0 |
| | [011] | 0 | 0 | 0 | 0 | 0 | 0 | 3 | 4 |
| | [111] | 0 | 0 | 0 | 0 | 0 | 0 | 0 | 29 |

Table 3. The confusion matrix obtained when the SVM network is used for fault classification

Predicted

| | | [000] | [100] | [010] | [001] | [110] | [101] | [011] | [111] |
|---|---|---|---|---|---|---|---|---|---|
| Actual | [000] | 39 | 0 | 0 | 0 | 0 | 0 | 0 | 0 |
| | [100] | 0 | 3 | 0 | 0 | 0 | 0 | 0 | 0 |
| | [010] | 0 | 0 | 3 | 0 | 0 | 0 | 0 | 0 |
| | [001] | 0 | 0 | 0 | 3 | 0 | 0 | 0 | 5 |
| | [110] | 0 | 0 | 0 | 0 | 3 | 0 | 0 | 0 |
| | [101] | 0 | 0 | 0 | 0 | 0 | 3 | 0 | 0 |
| | [011] | 0 | 0 | 0 | 0 | 0 | 0 | 3 | 1 |
| | [111] | 0 | 0 | 0 | 0 | 0 | 0 | 0 | 33 |

Table 4. The confusion matrix obtained when the GMM network is used for fault classification

Predicted

| Actual | | [000] | [100] | [010] | [001] | [110] | [101] | [011] | [111] |
|---|---|---|---|---|---|---|---|---|---|
| | [000] | 39 | 0 | 0 | 0 | 0 | 0 | 0 | 0 |
| | [100] | 0 | 3 | 0 | 0 | 0 | 0 | 0 | 0 |
| | [010] | 0 | 0 | 3 | 0 | 0 | 0 | 0 | 0 |
| | [001] | 0 | 0 | 0 | 3 | 0 | 0 | 0 | 1 |
| | [110] | 0 | 0 | 0 | 0 | 3 | 0 | 0 | 1 |
| | [101] | 0 | 0 | 0 | 0 | 0 | 3 | 0 | 0 |
| | [011] | 0 | 0 | 0 | 0 | 0 | 0 | 3 | 2 |
| | [111] | 0 | 0 | 0 | 0 | 0 | 0 | 0 | 35 |

## 8. CONCLUSIONS

In this paper GMM and SVM were introduced to classify faults in a population of cylindrical shells and compared to the MLP. The GMM was observed to give more accurate results than the SVM, which was in turn observed to give more accurate results than the MLP.